\documentclass[sigconf]{acmart}
\AtBeginDocument{%
  \providecommand\BibTeX{{%
    \normalfont B\kern-0.5em{\scshape i\kern-0.25em b}\kern-0.8em\TeX}}}

\acmConference[WWW '22]{World Wide Web Conference}{February 21-25, 2022}{France}
\acmBooktitle{WWW '22: World Wide Web Conference France}
\usepackage{xargs}
\usepackage[colorinlistoftodos,prependcaption]{todonotes}
\newcommandx{\sr}[3][1=]{\todo[linecolor=red,backgroundcolor=red!25,bordercolor=red,#1,inline]{sr: #2}}
\newcommandx{\sj}[3][1=]{\todo[linecolor=blue,backgroundcolor=blue!25,bordercolor=blue,#1,inline]{sy: #2}}
\usepackage{pgf-pie}

\usepackage{upquote}
\begin{document}

\title{CLUE: Contextualised Unified Explainable Learning of User Engagement in Video Lectures}

\author{Sujit Roy}
\email{sujit.roy@manchester.ac.uk}
\affiliation{%
\institution{The University of Manchester}
 \streetaddress{Booth St W}
 \city{Manchester}
 \country{United Kingdom}
 \postcode{M15 6PB}
}

\author{Gnaneswara Rao Gorle}
\affiliation{%
 \institution{Brainalive Research Pvt Ltd}
  \city{Kanpur}
 \country{India}}
\email{gnaneswara@braina.live}

\author{Vishal Gaur}
\affiliation{%
 \institution{Kone Corporation}
  \city{Espoo}
 \country{Finland}}
\email{vishal.gaur@kone.com}

\author{Haider Raza}
\affiliation{%
\institution{University of Essex}
 \city{Essex}
 \country{United Kingdom}}
 \email{h.raza@essex.ac.uk}

\author{Shoaib Jameel}
\affiliation{%
 \institution{University of Essex}
 \city{Essex}
 \country{United Kingdom}
}
\email{shoaib.jameel@essex.ac.uk}

%\renewcommand{\shortauthors}{Roy, et al.}


%
%Information collection
%\sr{We are trying to exploit the other information in the video to decode the engagement score which can help the instructor even to modify the video for vast engagement. Our proposed model can also help in characterising the users as per content information. Significance: Our model can be used to }
%

\begin{abstract}
Predicting contextualised engagement in videos is a long-standing problem that has been popularly attempted by exploiting the number of views or the associated likes using different computational methods. The recent decade has seen a boom in online learning resources, and during the pandemic, there has been an exponential rise of online teaching videos without much quality control. The quality of the content could be improved if the creators could get constructive feedback on their content. Employing an army of domain expert volunteers to provide feedback on the videos might not scale. As a result, there has been a steep rise in developing computational methods to predict a user engagement score that is indicative of some form of possible user engagement, i.e., to what level a user would tend to engage with the content. A drawback in current methods is that they model various features separately, in a cascaded approach, that is prone to error propagation. Besides, most of them do not provide crucial explanations on how the creator could improve their content. In this paper, we have proposed a new unified model, CLUE for the educational domain, which learns from the features extracted from freely available public online teaching videos and provides explainable feedback on the video along with a user engagement score. Given the complexity of the task, our unified framework employs different pre-trained models working together as an ensemble of classifiers. Our model exploits various multi-modal features to model the complexity of language, context agnostic information, textual emotion of the delivered content, animation, speaker's pitch and speech emotions. Under a transfer learning setup, the overall model, in the unified space, is fine-tuned for downstream applications. Our results show that our model can detect engagement reliably with 85\% accuracy and the explainability component helps give feedback to the creator where the content needs further improvement.

\end{abstract}




\maketitle

%The story in my mind for the Introduction section:
%1. What is the problem? What is the motivation?
%2. Why this problem is interesting and important?
%3. What do we know until now about the problem, existing research? Personalisation?
%4. What do we do and why our research is exciting?
%5. What and how our research will have an impact?

\section{Introduction}
%1. What is the problem? What is the motivation?
The ongoing pandemic has resulted in various teaching and research organizations resorting to online lectures. What is predominantly seen today is that different academicians across the globe are creating teaching materials and have started to share them online with students as Open Educational Resources (OERs) \cite{ehlers2018oer} such as Massive Online Open Courses (MOOCs) to boost the online learning. Users are now overwhelmed by the amount of data, for instance, on YouTube alone, searching for ``Deep Learning Lectures'' retrieves hundreds of results ranging from content created by various individuals to organisations worldwide. The ideal option for any user is to select the one that they could engage with reliably. While beliefs and biases \cite{white2013beliefs} do surround the choice of the videos, e.g., videos from a popular academician in a well-known organisation could be regarded as more interesting than others, this may not always be true. Simply relying on user ratings or modelling text extracted from the video recordings alone might not lead to desirable results because these features do not capture the overall quality of the content. Given the extensive and discrete number of educational materials, new and automated management ways are being devised where mostly cascaded models are used to predict the engagement score \cite{bulathwela2020predicting}. Regarding the OERs, this means to scale down the learners' efforts of finding the material without compromising its quality. Such objectives are usually accomplished after studying the personalization factor \cite{bulathwela2020truelearn} that is defined as contextual engagement which determines the extent of learners' context about a particular learning source.

Utilising user feedback (e.g., the video is too one-sided or monotonous) is becoming an important and time-sensitive challenge for successfully leveraging intelligent and user-centric systems in various applications. In the domain of online educational resources, it is imperative to provide timely feedback of user engagement over a population. The feedback can not only make it easier for content creators to create suitable videos as per the target audience, but also it will be more effective in the online teaching tools. There has been growing interest in the domain of contextualised engagement in OER \citeauthor{bulathwela2020predicting}. We argue that only a context agnostic model \cite{bulathwela2020predicting} cannot provide true user engagement, instead, we need an adaptive model, which utilises the features of a video lecture extensively and gives feedback for the engagement. Additionally, we put forward a crucial explainability approach, that can assimilate the information to improve the engagement with the target population.

%2. Why this problem is interesting and important?
The problem of automatically studying engagement is important because educators can create content that will optimize the engagement levels on their content. Besides, manual techniques, such as employing an army of domain-expert volunteers to view long videos and providing feedback is too time-consuming. In our model, the automated system provides feedback to an educator that the engagement quality of the teaching content is not faithful. While it can be argued that personalisation is more appropriate in such problem scenarios, personalisation exploits users' historical data and many users might not be willing to share their data such as their location, their click-through patterns, among others. Therefore, works such as \cite{bulathwela2020predicting} have largely focused on non-personalised prediction problems. What is interesting in our work is that we provide feedback to the content creator to help understand which areas they need to improve, e.g., should they add more animations. One key advantage of our model is that it is crucial to a user who might end up wasting plenty of time searching for ideal content suited to their learning behaviour, e.g., some users prefer more animations in their videos.

%3. What do we know until now about the problem, existing research? Personalisation?
There has been a growing interest in the contextualised engagement \cite{bulathwela2020sum} in recent years which model text and likes associated with videos. We argue that there are other features of the video delivery mechanism, which can increase or decrease user engagement. The categorical model assumes that there are discrete emotional categories such as Ekman’s six basic emotions -- anger, disgust, fear, joy, sadness, and surprise \cite{ekman1992there}. Emotion recognition has been explored considerably \cite{li2016hybrid, wang2016multi,  baziotis2018ntua, meisheri2018tcs, du2018mutux, alswaidan2020survey}. In our model, we have modelled the emotional variation of the speaker to time. There have been previous attempts at predicting emotions from speech \cite{lugger2009combining, schuller2005robust, nicholson2000emotion, engberg1996documentation, burkhardt2005database, nwe2003speech} where the average classification accuracy of speaker-independent speech emotion recognition systems is less than 70\% in most of these proposed techniques. For example, it is 50\% in \cite{nicholson2000emotion}, 67\% in \cite{engberg1996documentation}, 80\% in \cite{burkhardt2005database}, and 65\% in \cite{nwe2003speech}. We go beyond these existing methods and develop our speaker-independent emotion recognition model unified with an ensemble of learners to model the user engagement problem reliably.

%4. What do we do and why our research is exciting?
Our hypothesis is that engagement cannot be directly exploited using text alone in lecture videos as it has been done in some of the past works \cite{bulathwela2020predicting}. There is various complementary information that we could exploit from online lecture videos such as speaker intonation, speaker's use of animations, emotions, among others that have not been explored in prior works. We develop a novel unified framework that goes beyond the current techniques that measure user engagement. We expect that over time, our research would have a significant impact in the education domain, where users could find a plethora of engaging free materials online. It will lead to a significant positive step towards achieving the global development goals. To the best of our knowledge, this is the first work in the area of contextualized engagement, where we exploit other features beyond just text extracted from video lectures.

Our key contributions include:
\begin{itemize}
    \item A novel unified model that learns to fine-tune its parameters by exploiting the predictive error of several pre-trained models in an ensemble setup.
    \item The model not only predicts an engagement score but also provides explainable feedback to the content creator.
    \item We have conducted experiments on publicly available free videos and the dataset will be shared with the community given the lack of freely available public benchmark datasets in this problem domain.
    \item Given the importance of low-resource data modelling, we demonstrate that our model is reliable under low-resource settings, i.e., settings when we have fewer data.
\end{itemize}

\begin{figure*}
    \centering
    \includegraphics[scale=0.28]{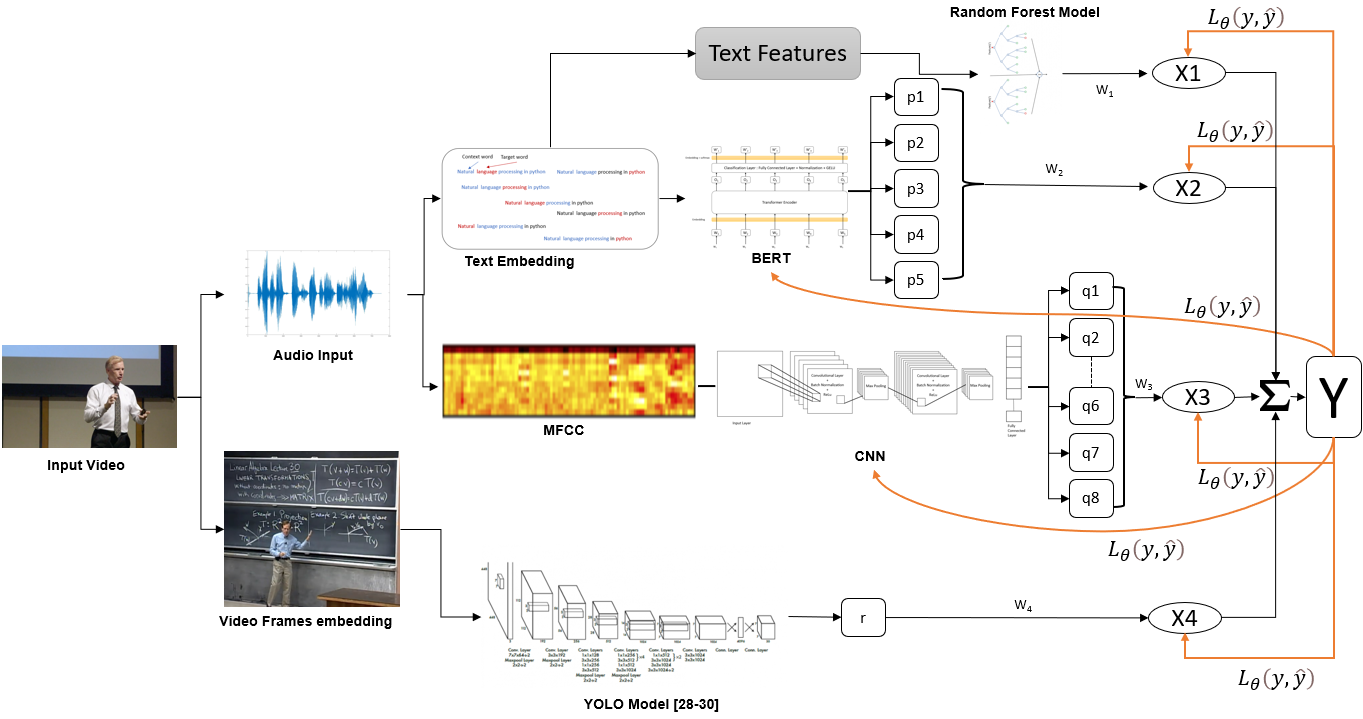}
    \caption{Architecture diagram of our unified CLUE model.}
    \label{fig:model}
\end{figure*}

\section{Related Work}
\citeauthor{bulathwela2020predicting} in \cite{bulathwela2020predicting} developed a technique to learn text feature vectors from video lectures. They obtained a proprietary dataset from VideoLecture.net, which is not publicly available. They extracted different features from the videos in this dataset and trained a regression model to determine the overall prediction score. While the work is closely related to ours, there are some key differences. We have developed a novel unified approach to model various multi-modalities present in the videos which \cite{bulathwela2020predicting} have not done. We have argued that only textual features might not suffice when predicting the overall engagement score. While our results cannot be directly compared with theirs because of the non-availability of their dataset, our model is superior in a variety of ways, mainly in how it models the problem in a unified setup modelling different multi-modalities including modelling explainability. \citeauthor{zhou2021modeling} in \cite{zhou2021modeling} developed a novel unsupervised model sequence mining and information retrieval coupled with a clustering algorithm to extract engagement patterns of learners. Their goal is to mainly extract consistent patterns in learning behaviour. As a result, this work is fundamentally different from ours in various ways, where our goal is not to learn consistency levels.

\citeauthor{chen2020automated} in \cite{chen2020automated} have developed a technique to model automated disengagement that detects learners' maladaptive behaviours, e.g. mind-wandering and impetuous responding. While their work does not develop a novel computational model as ours, our framework is both novel and fundamentally different from this work where our goal is to model user engagement and provide feedback to the user through its explainability model. Recently, \citeauthor{bulathwela2020truelearn} in \cite{bulathwela2020truelearn} developed a novel recommendation framework that considers several background information of a user, for instance, learner's knowledge of the topic. While they extend their prior work in \cite{bulathwela2020predicting}, their main focus in this work is to develop a recommender system for predicting the engagement score as in our work. In \citeauthor{bulathwela2020vlengagement} \cite{bulathwela2020vlengagement}, the authors have created a new dataset of video lectures. However, they have not shared the videos, hence other researchers are unable to extract new features from them.

We have exploited different features, which could help to improve the engagement score of our model, for instance, we have exploited emotions from text and speech to model engagement. We have also exploited object detection techniques to further improve our results. As a result, our model can learn from various key representative features which are crucial towards determining the overall engagement score including modelling explainability. For instance, by modelling different objects in the videos during teaching, we can model that the teacher is using different teaching methods. The importance of emotions has been highlighted in many other works, for instance, \citeauthor{nias1996thinking, nias1996thinking} studied the emotional aspects of teachers in the UK. The authors conclude that the emotions of the instructor play a vital role in teaching since it helps improve engagement of the subject. Recently, in \cite{jimenez2021scientific}, \citeauthor{jimenez2021scientific} studied the important question, ``What is the nature of preservice teachers' emotions throughout their engagement in the sequence?'' They concluded that ``emotions in science education as they illustrate the importance of providing preservice teachers with opportunities to explore their emotions especially in relation to self-regulation when engaging in teaching sequences in teacher preparation.''

\section{CLUE: Our Novel Framework}
In this section, we describe our full framework which models the engagement score along with providing explanations to help the content creator improve their videos. The engagement score measures how likely the user will engage with the content. The explanations provided by the model will help the content creator in understanding the key shortcomings in the content which will help improve the quality. The key design principle of our model is to exploit the advantages of different existing pre-trained complementary models. These models are combined, as an ensemble of models, working as a unified machine learning model where they make predictions jointly rather than as a cascaded model which will result in error propagation. This design paradigm gives us a direct advantage that we can model multi-modal features using the most suitable computational model for the feature-type, for instance, frame representations can most ideally be learned using a Convolutional Neural Network (CNN) model than a random forest model which is most ideally designed to model text in our problem setup, and subsequently make predictions in a unified way. While all individual pre-trained models play a key role in the overall predictive performance, some can be further fine-tuned based on the data characteristic leading to more faithful results, for instance, the pre-trained BERT \cite{devlin2018bert} model can be fine-tuned given the data than using the original fine-tuned BERT alone. Similarly, we can fine-tune the CNN model on our data in an iterative way. As a result, we exploit the key advantage of transfer learning.

We model emotion-based features from text and speech followed by object detection. Emotion in teaching has received plenty of attention in literature \cite{nias1996thinking}. Emotions in the classroom are not only a private matter but also a political space in which students and teachers interact with implications in larger political and cultural struggles \cite{zembylas2007power}. Besides, automatic object detection can help the model understand what else a teacher uses while teaching, e.g., are there some classroom activities organised. In our model, these complementary models contribute towards our overall goal of engagement prediction and explainability. To extract and learn different features from publicly available datasets, we have used a variety of state-of-the-art models which we apply in the predictive analysis tasks. Our overall framework comprises of, 1) text-based context agnostics, 2) text-based emotions, 3) speech-based emotions and 4) object detection, which we show to play a significant role towards our overall goal. There are certain key tasks that we need to do, for instance, extraction of features, learning those features using relevant models and using our unified machinery on these individual complementary pre-trained models to derive their weights leading to a prediction score.

In our novel modelling architecture depicted in Figure~\ref{fig:model}, from the text transcript, we predict \(X1\) by random forest and \(X2\) by BERT model, where \(P1\), \(P2\), \(P3\), \(P4\), \(P5\) is the output of text-based emotion. Using the audio feature we predict \(X3\) based on the probability of \(Q1\), \(Q2\), ..., \(Q8\) which represents speech-based emotion. Finally, \(X4\) is the count of objects that appeared in the video. We then combine these output of models and predict \(Y\) where the parameter training of our model and the fine-tuning of the pre-trained BERT and CNN models take place simultaneously based on the feedback of $\hat{Y}$.

\begin{table*}
\centering
\large
\caption{Extracted features from the VLN data-set}
\resizebox{\textwidth}{!}{
\begin{tabular}{|ll|}
\hline
\textbf{Feature} & \textbf{}  \\ \hline
Conjugate\_rate & Count of conjunctions used by the speaker in a particular video \\ \hline
Pronoun\_rate & Count of pronouns used by the speaker in a particular video. \\ \hline
preposition\_rate & Count of prepositions used by the speaker in a particular video. \\ \hline

tobe\_verb\_rate & Count of times the speaker used (``be'', ``being'', ``was'', ``were'', ``been'', ``are'', ``is'') in a video lecture. \\\hline
 
auxiliary\_rate & Count of times speakers used auxiliary verbs in a video. \\ 
 \hline
normalization\_rate & Count of words used by the speaker which were  ended with suffixes (``tion'', ``ment'', ``ence'', ``ance'')\\ \hline
 
fraction\_stopword\_coverage & It is the ratio of stopwords used by the speaker to the total stopwords \\ \hline
 
Fraction\_stopword\_presence & It is the ratio of stopwords used by the speaker to the total number of words used by the speaker in a particular video \\\hline
 
Easiness & It will give the readability level of the text of a video. \\ \hline
 
Document\_entrophy & Entropy of words for a particular video.\\ \hline
 
word\_count & Total count of words used by a speaker in a particular video. \\\hline 
 
title\_word\_count & Count of title words for a particular video. \\ \hline
 
Duration & Length (Time) of video file. \\\hline
 
Speaker speed & Count of words used by a speaker per minute.\\
\hline
\end{tabular}%
}
\label{table1}
\end{table*}
\subsubsection{Text based Context Agnostics}

We have used VideoLecturesNet (VLN) dataset \cite{bulathwela2020predicting} which has been categorized into 21 different subjects, such as Computer Science, Physics, Philosophy, etc. To extract the context engagement model we extracted features from the VLN dataset, which are, `duration', `conjugate rate', `normalization rate', `tobe verb rate', `auxiliary rate', `preposition rate', `pronoun rate', `document entropy', `easiness', `fraction stopword coverage', `fraction stopword presence', `title word count', `word count', `speaker speed', `median engagement rate'. The subset of cross-modal and language-based features were selected from the VLN dataset. The 14 extracted features can be seen in Table~\ref{table1}. The description of the features is also shown in table \ref{table1} which are our novel features to suit our problem task. We split the data in 67\% for training and 33\% for the test. The random forest regressor was used to train as the model with median engagement rate as a prediction variable ranging from 0-1 and was stored as X1 in CLUE. This model gave us better results in our case than the support vector-based model. Mean squared error was calculated from the predicted variable. The median engagement rate was calculated based on user feedback, star rating of videos, number of views, and likes for the videos.

\subsubsection{Text based Emotions}

For evaluating emotions in a text we have used the International Survey on Emotion Antecedents and Reactions (ISEAR) and Dailydialogue publicly available datasets. The ISEAR dataset \cite{scherer1994evidence} contains the emotional statement that helped us to further train the model using the textual data. It contains 7666 sentences which are further divided into 6 emotional categories. 

The dailydialogue dataset \cite{li2017dailydialog} is a high-quality multi-turn dialog dataset. The dataset is constituted of human written statements which makes them less noisy. The dataset contains information that reflects our daily conversations and consists of a variety of topics about our daily lives. The dataset is also manually annotated in similar emotional categories to the aforementioned datasets. This dataset contains a total of 13,118 dialogues which are then split into 11,118 training dialogues and 1000 dialogues of validation and test set each.

The emotions were characterised into 5 categories, i.e. Joy, sadness, anger, fear, and neutral. The training size of the dataset was 27,261 and validation was 3,393. The model used for training the dataset was K-train based text BERT model. The maximum length of the unigrams is 35000 and the tokens are 350. We run the model for 5 epochs with a batch size equal to 12. The learning rate was kept $2e^{-5}$. The output of the model was treated as an array with the probability of all the emotions, shown as p1, p2, p3, p4, p5 and was stored as X2 in CLUE.

\subsubsection{Speech based Emotions}
Emotions play an important role in teaching \cite{sutton2003teachers}. A monotonous video without any emotions will be relatively less engaging than those videos where the teacher exploits emotions. Besides emotion detection in the text, we also conduct emotion detection in speech which would lead to a more reliable understanding of the engagement factors in videos.
\begin{table}
\centering
\small
\caption {Parameters for Speech based emotion detection model. B is referred to as batch size.}
\begin{tabular}{|l|l|l|l|}
\hline 
Layers          & Output Shape  & Param   & Activation    \\\hline
Input     &  {[}B, 180{]}   &    &  \\ \hline
conv1D\_1     & {[}B, 161, 128{]}      & 2688 & ReLU\\\hline
batch normalisation     & {[}B, 161, 128{]}      & 512 & \\\hline
conv1D\_2     & {[}B, 152, 64{]}      & 81984 & ReLU\\\hline
batch normalisation     & {[}B, 152, 64{]}      & 256 &  \\\hline
flatten   & {[}B,4864{]}      &  & \\ \hline
dense\_1   & {[}B,520{]}      & 5059080 & \\ \hline
dense\_2   & {[}B,8{]}      & 4168 & Softmax \\ \hline
\multicolumn{4}{|l|}{Total params : 5,148,688}        \\ \hline
\multicolumn{4}{|l|}{Trainable params : 5,148,304}\\\hline
\multicolumn{4}{|l|}{Non-trainable params : 384}\\\hline
\end{tabular}
\label{table2}
\end{table}

To calculate Speech-based emotions three features, i.e., Mel-frequency cepstral coefficients (MFCC), chroma, and Mel spectrogram frequency were extracted from the speech waveform from the RAVDESS dataset. Ryerson Audio-Visual Database for Emotional Speech and Song (RAVDESS) dataset \cite{livingstone2018ryerson} consists of speech and song, audio and video files. For our analysis, we focused on the emotional speech and song files. There are 1440 files in the RAVDESS dataset that assist in analyzing the emotions from the speech. The RAVDESS dataset contains 24 actors (12 male and 12 female), who record the speech in lexically similar statements in a neutral North American accent. The speech dataset is further categorized into seven different emotions namely, ``Calm'', ``Happy'', ``Sad'', ``Angry'', ``Fearful'', ``Surprise'' and ``Disgust''. There are mainly two different levels of emotional intensities (Normal and Strong). There is also an additional third neutral expression.

The speech was classified into 8 categories, which are, ``Angry'', ``Sad'', ``Happy'', ``Neutral'', ``Fear'', ``Disgust'', ``Surprise'' and ``Calm''. MFCC:  Mel Frequency Cepstrum (MFC) is a representation of linear cosine transform of a short-term log power spectrum of a speech signal on a non-linear Mel scale of frequency. Mel-frequency cepstral coefficients (MFCCs) are together make up an MFC. MFCC extraction is of the type where all the characteristics of the speech signal are concentrated in the first few coefficients \cite{muda2010voice}. Chroma-based features are used for identifying pitch-based information or they can also be referred to as pitch class profiles. The chroma representation is used for intensities of 12 distinct musical chromas of the octave at each time frame. By using chroma we generated a chromagram based on 12 pitch classes. The pitch classes in the particular order are as follows C, C\#, D, D\#, E, F, F\#, G, G\#, A, A\#, B.

The total number of training samples was 1152 and validation samples were 288. We used 1D CNN for training the data. For the input layer, we extracted all the features and combined them horizontally, where the length of the input vector was 180. The length of the MFCC feature was 40, Mel was 128 and chroma was 12. After the input layer, we used 1D CNN layer with 128 filter sizes of 20 and a stride of 1. Post the CNN layer we used the batch normalization layer and the activation function was ReLU. We used another layer of 1D CNN with 64 filter sizes of 10. Another batch normalisation layer was used and the activation function was ReLU. We further used a dense layer of size 520 and then another dense layer of size 8 with softmax as an activation function. The training batch size was 16, the learning rate was $10^-4$, the loss was categorical cross-entropy and Adam was used as a training optimizer. Table~\ref{table2} describes the model architecture and parameters. The output of the model was treated as an array with the probability of all the speech-based emotions, shown as \(q1\), \(q2\),$\cdot \cdot \cdot$, \(q8\) and was stored as \(X3\) in CLUE.

\subsubsection{Object Detection}
Our motivation to include object detection is primarily to capture different objects in the videos \cite{sieber2001teaching, kay2008exploring}, for instance, if there are animations in the videos, it would result in more objects than just the teacher and the students. Other use cases include if the teacher uses a variety of objects in a class to teach students in addition to traditional objects already found in the classrooms, for instance, a Physics teacher using a range of real-world objects to explain a concept. YOLO v3 uses logistic regression to compute the target score. It gives the score for all targets in each boundary box. YOLO v3 can give the multilabel classification because it uses a logistic classifier for each class in place of the softmax layer used in YOLO v2. YOLO v3 uses darknet 53. It has fifty-three layers of convolution. These layers are more in-depth compared to darknet 19 used in YOLO v2. Darknet-53 contains mainly 3x3 and 1x1 filters along with bypass links \cite{redmon1804yolov3, redmon2015you, redmon2018yolov3}. The output of the model was r where it counted the objects and animations that appeared in the video with respect to time and was stored as X4 in CLUE.

\subsection{CLUE: Our Unified Engagement Score Model}
We have used four different pre-trained models, giving complementary knowledge, to decode different features present in video lectures. While these models are pre-trained on specific tasks, our goal is to exploit these model parameters and develop an end-to-end novel unified deep learning framework. Out of these pre-trained models, CNN and BERT can be simultaneously fine-tuned while the unified model training is underway. In fine-tuning, we freeze some of the layers and fine-tune only specific layers which are needed for our task, for instance, in the pre-trained text language model, we only fine-tune the contextual layers, mainly, layer 12. Our framework is depicted in Figure~\ref{fig:model} where we extract audio from video, and audio extraction of the speech to text is performed using the IBM Watson speech to text platform. After a speech to text, we have extracted 13 features based on their continued use in studies \cite{dalip2011automatic, warncke2013tell, ntoulas2006detecting, guo2014video, bulathwela2020predicting}. Table \ref{table1} depicts these extracted features, and apart from these features, we also trained the model to extract emotions from the text data. For training, the Emotion ISEAR, DAILYDIALOG, and KAGGLE Datasets were used. The model was trained on 27,261 sentences considering five classes, namely, ``Joy'', ``Sad'', ``Fear'', ``Anger'', ``Neutral''. For emotion classification, the BERT model has been used. Mathematically, our model formulation is shown in Equation~\ref{eq:eq_linearcombin} where we linearly combine (convex combination) different models using four parameters. While these parameters could be arbitrarily assigned or given the same weights, we have trained these model parameters based on the data. To this end, we have used the backpropagation model to update these model parameters in each iteration, and simultaneously fine-tuned the individual models.

\begin{figure}
	\centering
	\includegraphics[width=0.48\textwidth,keepaspectratio]{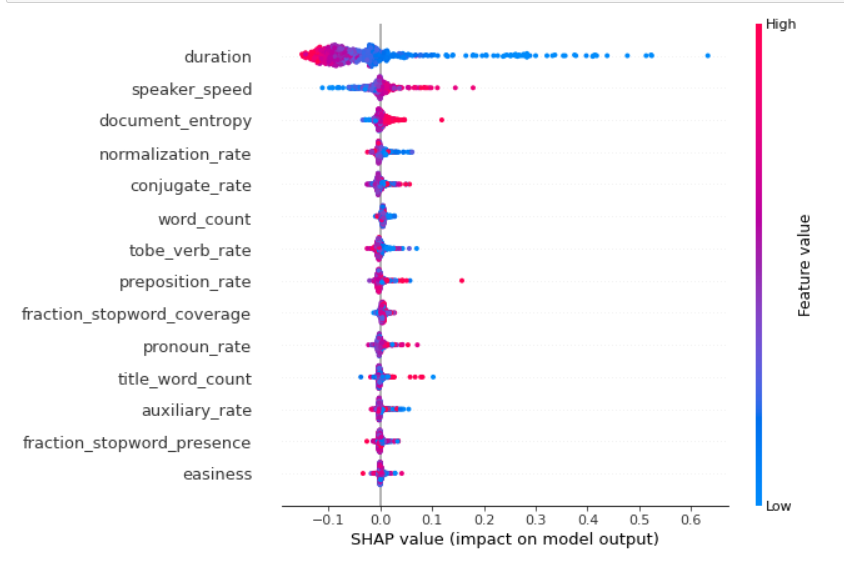}
	\caption{Shap values for VLN dataset for context engagement. The duration of video is a very important feature and it is showing that longer length of video shows drop in engagement.}
	\label{fig:vlnshap}
\end{figure}

\begin{figure}
	\centering
	\includegraphics[width=0.435\textwidth,keepaspectratio]{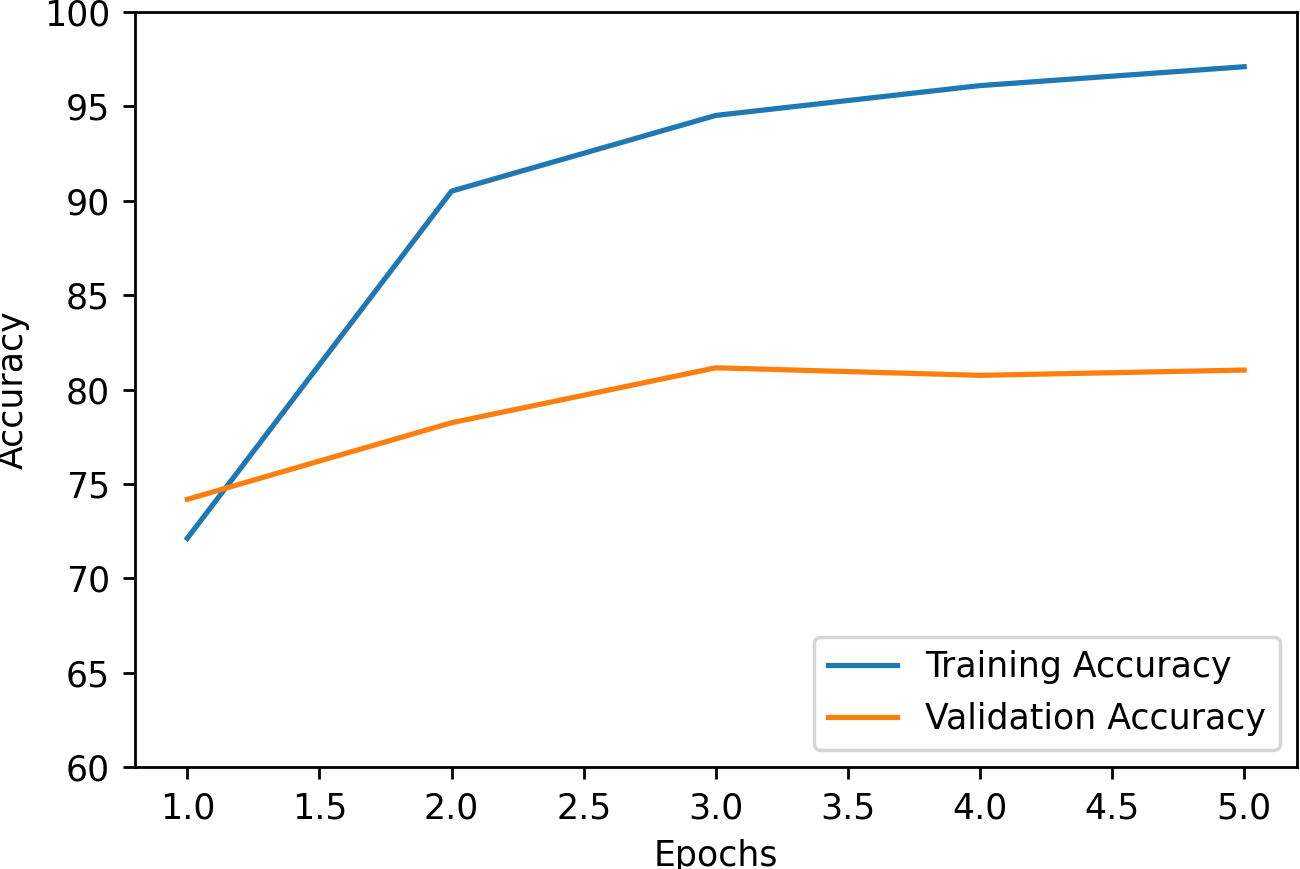}
	\caption{Training and test accuracy for decoding text based emotion using K-train BERT architecture.}
	\label{fig:bertaccuracy}
\end{figure}

In Equation~\ref{eq:eq_linearcombin}, \(X_1\) is based on contextual engagement provided with contextual engagement score ranging from 0-1 based on 14 textual features. This is the regression-based model. \(X_2\) provides the overall emotional distribution over 5 classes for the lecture transcript. This is the text-based model which exploits representation vectors obtained from the BERT language model. \(X_3\) provides the emotion feedback with reference to time, based on the speaker's tone and delivery speech which mainly exploits the speech data. \(X_4\) detects the number of animation and objects with reference to time, which is the object detection model. 

\begin{equation} \label{eq:eq_linearcombin}
    y = \alpha X_1 + \beta X_2 + \gamma X_3 + \delta X_4
\end{equation}

\noindent where \(y\) is the prediction score. The individual weight parameters $\alpha, \beta, \gamma, \delta$ are the coefficients. Initially, $\alpha, \beta, \gamma, \delta$ will be initialised with random weights. After every watched video, we are collecting the user rating for the video (out of 5) and positive and negative comments. This user feedback is the real truth and is denoted by $\hat{y}$. To minimise the error we use the Huber loss \cite{huber1965robust} given by:

\begin{equation}
    L_{\theta}(y, \hat{y}) = 
 \begin{cases} 
 0.5 * (y - \hat{y})^2, \quad |y - \hat{y}| \leq \theta \\ 
 \theta * (|y - \hat{y}| - 0.5 * \theta),   \quad \text{otherwise}
 \end{cases} 
\end{equation}

\noindent where $\hat{y}$ is the normalised user feedback truth, $y$ is the predicted output and $\theta$ is the hyperparameter for a large or small error. Our objective is to minimize $L_{\theta}(y, \hat{y})$ based on $\alpha, \beta, \gamma, \delta$. The loss will be backpropagated for the individual model as well as the coefficients. Since it is a linear equation, the coefficients can also be optimised by the Gaussian process, but it would be highly unlikely that there will be a unique solution. Therefore, by using backpropagation we are aiming to optimise the model as well as the coefficients for the prediction of engagements. We run the backpropagation model for a certain number of iterations until the model parameters converge. In the end, we obtain the converged weight parameters followed by fine-tuned models learned in a unified way. The advantage that we get is that these parameters are trained in a consolidated parameter space which leads to more reliable results than approaches that use cascaded techniques where the output of one or more is fed as input to the next model. We found that, in our case, the results obtained from cascading models were too poor and that it was difficult to engineer the pipeline framework because there is no such pre-defined sequence rule for the models that we could engineer, e.g., should \(X_1\) come before \(X_2\) or vice-versa. In contrast, in our CLUE, all models work simultaneously. For initialising, the coefficients are $\alpha=0.5$, $\beta=0.1$, $\gamma= 0.2$ and $\delta=0.2$.

\begin{figure}[h]
	\centering
	\includegraphics[width=0.5\textwidth,keepaspectratio]{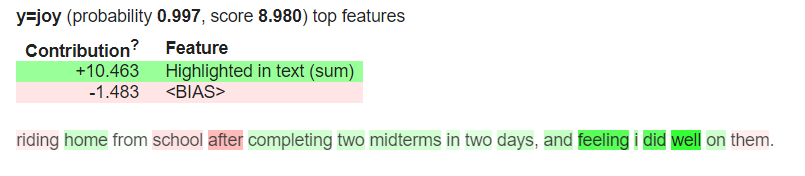}
	\caption{Explainability for the Ktrain BERT model, where green indicates the positive contribution and red negative.}
	\label{fig:bertexplain}
\end{figure}

\section{Experiments and Results}
We have presented our experiments in two units where we first report results from the individual models to showcase that they contribute reliably. We then present results from our overall architecture to showcase how these pre-trained models contribute in a unified way when used under a transfer learning setup. Note that individual model results are treated as baseline results demonstrating that individually these models are not suitable to address the problem reliably.

\subsection{Individual Model Performance}
Using random forest regressor for contextualised engagement we obtained a mean squared error of 0.0173. Figure \ref{fig:vlnshap} shows the shap summary plot of different features for deciding by random forest regressor. The summary plot combines feature importance with feature effects. Each point on the summary plot is a Shapley value for a feature and an instance. The position on the y-axis is determined by the feature and on the x-axis by the Shapley value. The colour represents the value of the feature from low to high. The features are ordered according to their importance. We can see that the length of the video is the most important feature and longer lengths of videos have less impact on the model. Preposition\_rate has a low effect on the model. Similarly, tobe\_verb\_rate has major effect on model but the used rate should be <0.03\% (``be'', ``being'', ``was'', ``were'', ``been'', ``are'', ``is''). Auxilary\_rate (``will'', ``shall'', ``cannot'', ``may'', ``need to'', ``would'', ``should'', ``could'', ``might'', ``must'', ``ought'', ``ought to'',``can't'', ``can'') will provide with positive effect if the used rate is <0.025\%. Speaker speed has more importance to the model and speed contributing positively to engagement is 115-120 words per minute. It should not be very low or very high. The easiness level determined should be more than 83 to have a positive impact on the model. A normalisation rate greater than 0.1 has a positive effect on the engagement score. Preposition\_rate has a very minor effect on the model.

Figure \ref{fig:bertaccuracy} shows the training and validation accuracy for the k-train based BERT model for decoding text-based emotion. We used a pre-trained k-train model and further trained it on our dataset. The model was provided with the text extracted from speech to text. The model runs for 5 epochs and post that no improvement was seen and the training was stopped. The model achieved a test accuracy of 81\% with 5 classes. Table \ref{table3} shows the precision, recall, and F1 score of individual emotions. Precision for Joy and fear is the same 0.85\%, anger is 0.82\%, neutral is 0.79\% and sadness is 0.77\%. 
Macro-average values is calculated by:
\begin{equation}
    P = \frac{P_1+P_2 + \cdot+ P_n}{N} 
\end{equation}
where \(P\) can be precision or recall or F1 score. $P_i$ is precision/recall/f1 score for class \(i\) and \(N\) is the total number of class.

\begin{table}
\centering
\caption {Results for Emotion classification using Ktrain-BERT Model for text} 
\label{table3}
\begin{tabular}{|l|l|l|l|}
\hline 
Emotion          & Precision  & Recall   & F1-score    \\\hline
Joy & 0.85  & 0.84  & 0.84 \\ \hline
sadness & 0.77  & 0.83  & 0.80 \\\hline
fear & 0.85 & 0.83  & 0.84\\\hline
anger & 0.82    & 0.75  & 0.78\\ \hline
neutral & 0.79  & 0.83  & 0.81 \\ \hline
macro avg & 0.82    &  0.82   &   0.81    \\ \hline

\multicolumn{4}{|l|}{accuracy : 0.81}  \\ \hline
\end{tabular}
\end{table}

\begin{figure}
	\centering
	\includegraphics[width=0.43\textwidth,keepaspectratio]{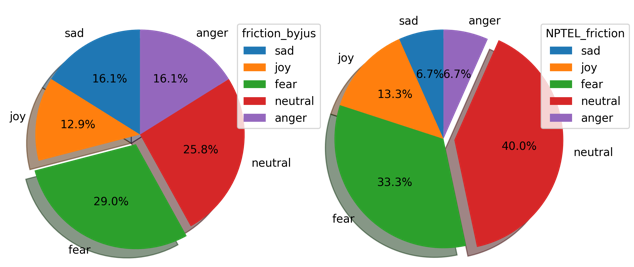}
	\caption{Emotion distribution of two different video by different open content providers on the same topic.}
	\label{fig:emote}
\end{figure}

Figure \ref{fig:bertexplain} shows the model explainability behind the emotion ``Joy'', where green shows the positive contribution and red shows negative emotion towards ``Joy''. Figure \ref{fig:emote} shows the overall distribution of emotions in the text data. It was observed that the content with more varying emotions has higher engagement than the content with one emotion.

\begin{figure}
	\centering
	\includegraphics[width=0.8\columnwidth,keepaspectratio]{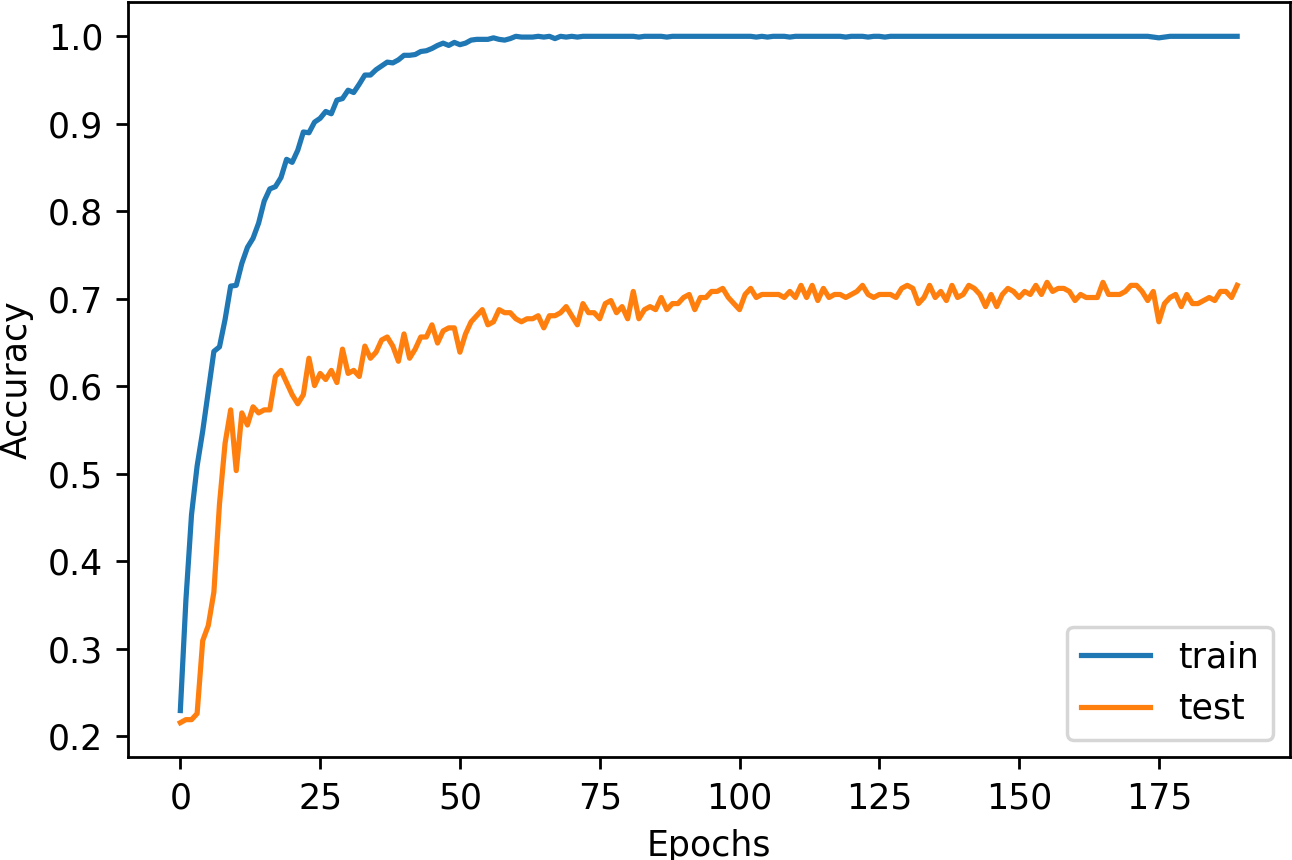}
	\caption{Training and test accuracy for decoding speech based emotion from Ravdess dataset.}
	\label{fig:trainingspeech}
\end{figure}

Figure \ref{fig:trainingspeech} shows the training and test accuracy for speech-based emotion detection. The model was trained for 190 epochs and achieved an overall accuracy of 70.83\% with 8 classes. 

\begin{figure}
	\centering
	\includegraphics[width=\columnwidth,keepaspectratio]{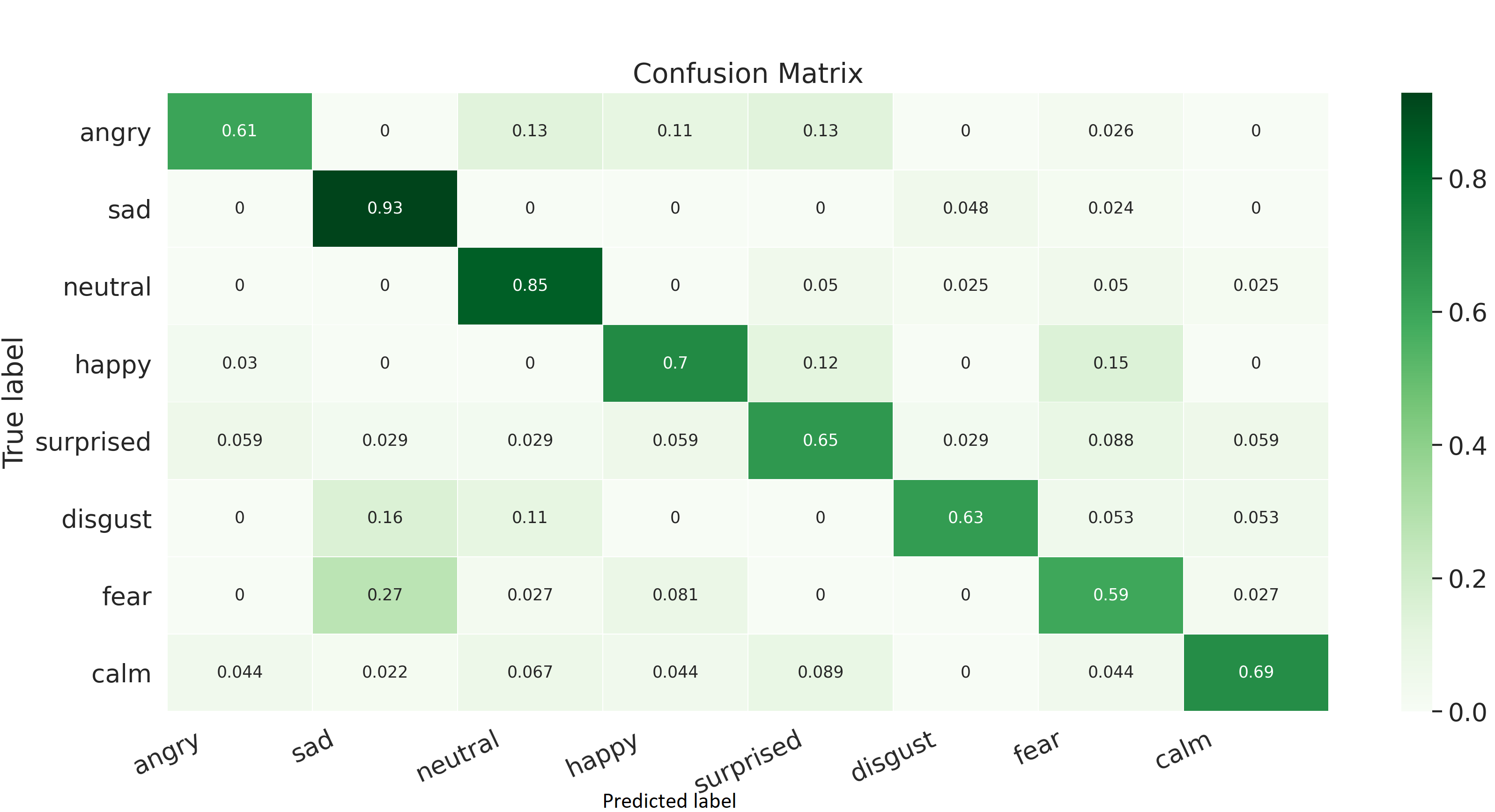}
	\caption{Confusion matrix for speech based emotion detection}
	\label{fig:cms}
\end{figure}

Figure \ref{fig:cms} shows the confusion matrix of emotions from speech data. The precision for angry is 0.82 \%, calm is 0.72\%, disgust is 0.74\%, fearful is 0.68\%, happy is 0.59\%, neutral is 0.75\%, sad is 0.59\% ans surprised is 0.86\%. Macro average precision, recall and f1 score is 0.72\%, 0.71\% and 0.71\% respectively.

\begin{figure}
	\centering
	\includegraphics[width=0.43\textwidth,keepaspectratio]{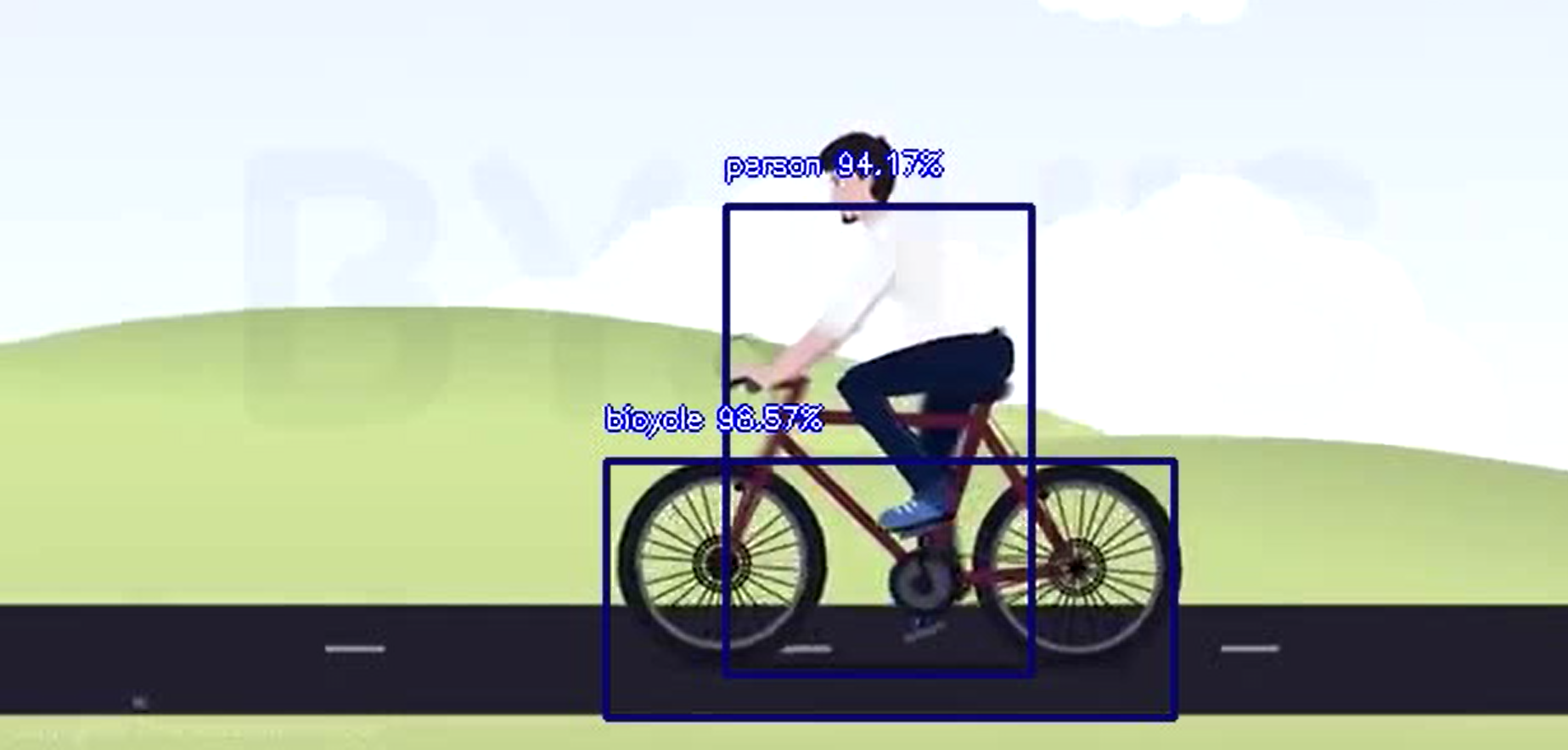}
	\caption{Yolov3 object detection in online video tutorials.}
	\label{fig:yolo}
\end{figure}

Figure \ref{fig:yolo} shows the detection of objects inside the specific video frame with 3fps and it keeps the count of the objects shown with reference to time. As a result, we can automatically detect the activities with different objects in the video. We are extracting the information of objects and matching it with the pretext context of that object for the relevancy of the topic. In this way, we can gauge the role of various objects towards the overall engagement score.

\subsection{CLUE Results}
We have used our proposed architecture CLUE on our new dataset, where we used the user ratings based on engagement for the video as the target variable and benchmark. For collecting this dataset, we have deployed the model on the cloud server, and users are shown a video from open-source videos lectures according to their interest. After completing the video users are asked to provide a rating for their engagement on a scale of 1 to 10. We have, until now, 50 videos in our collection dataset. The average video length was 29.5 minutes, and the topics cover domains of school level physics, literature and history. Videos were obtained from open source lectures, such as NPTEL, byjus, and Unacademy. Note that even with this dataset size, we can obtain reliable features from these full-length videos to train our model. We expect that with the large-scale datasets such as those used in \cite{bulathwela2020predicting}, we can further help improve our model performance.

\begin{figure}
	\centering
	\includegraphics[width=0.39\textwidth,keepaspectratio]{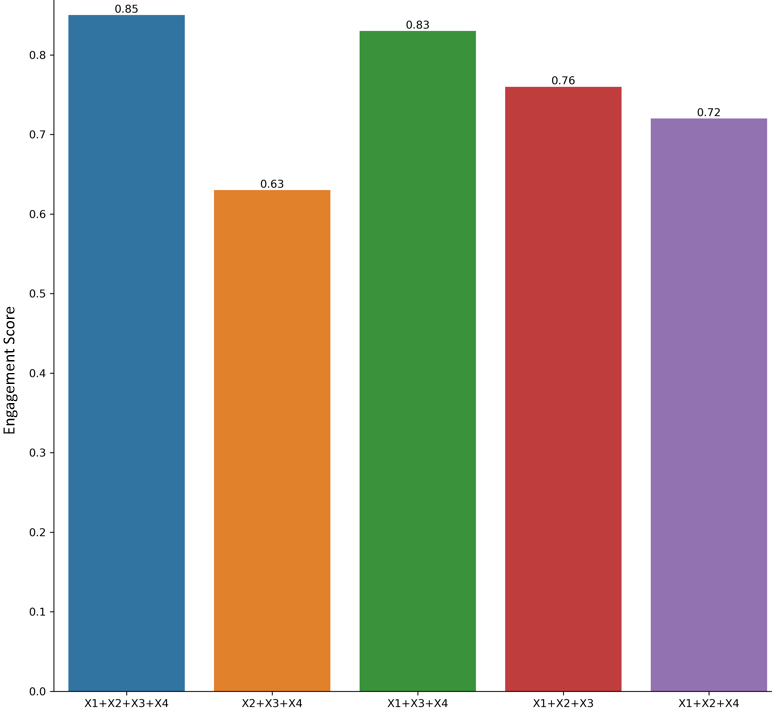}
	\caption{Variation in predicted mean engagement score by leaving one model out. We can notice that compared to individual baseline models mentioned above and leaving out certain models out, our full model obtains the best result quantitatively.}
	\label{fig:score}
\end{figure}

For every video processing, it was separated in two segments, one with video frame embeddings and the other with audio. We extracted the features from the audio and video as discussed for CLUE and calculated the engagement score based on our pre-trained model. The ground truth mean engagement score as reported by users after normalisation is 0.88. The predicted mean engagement score by CLUE is 0.85. Figure \ref{fig:score} depicts the predicted engagement score by all the models and also the predicted engagement score if we leave one model out. Predicted engagement score $X_1 + X_2 + X_3 + X_4$ = 0.85, $X_2 + X_3 + X_4$ = 0.63, $X_1+ X_3+ X_4$ = 0.72, $X_1 + X_2 + X_4$= 0.76, $X1 + X_2 + X_3$ = 0.72. The engagement score  shows the impact of individual model on the final output.

\begin{figure}
	\centering
	\includegraphics[width=0.43\textwidth,keepaspectratio]{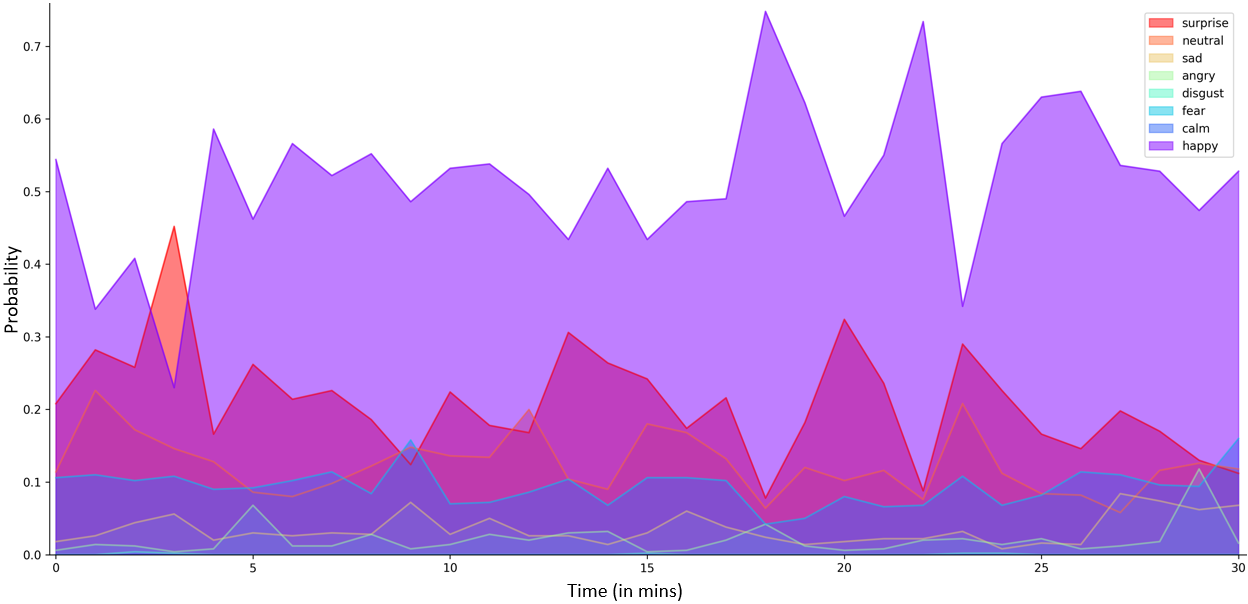}
	\caption{Speech emotion response of video 1 with respect to video time stamp.}
	\label{fig:ae1}
\end{figure}

\begin{figure}
	\centering
	\includegraphics[width=0.43\textwidth,keepaspectratio]{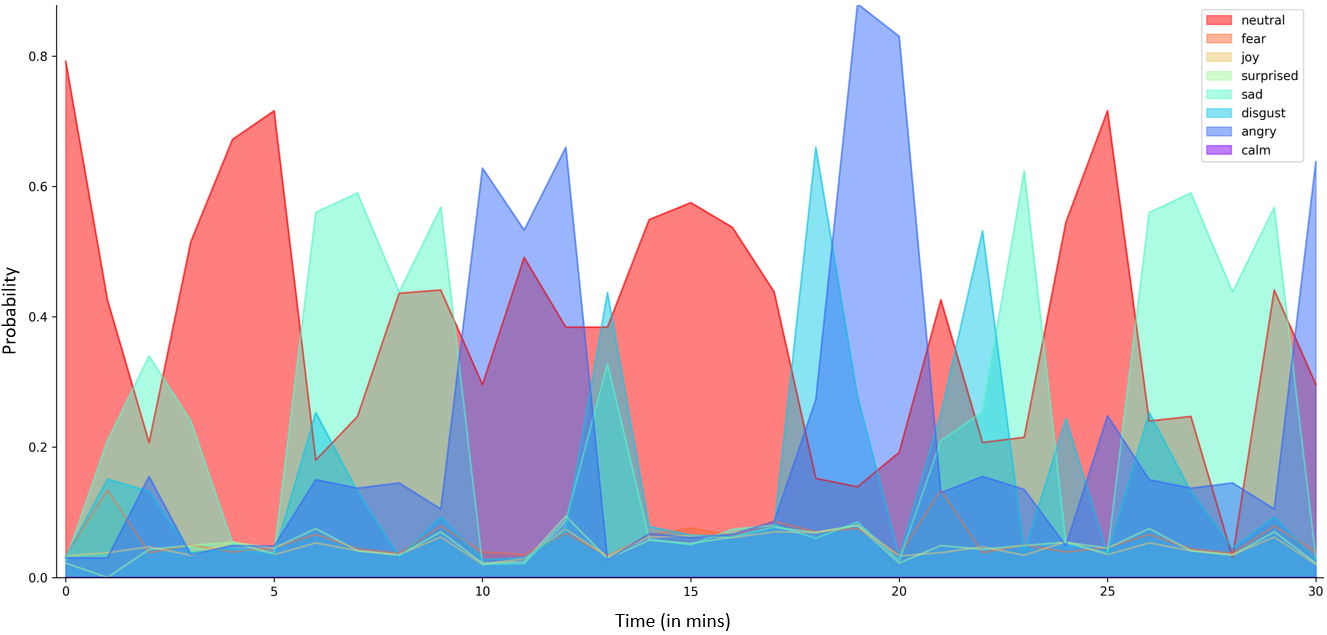}
	\caption{Speech emotion response of video 2 with respect to video time stamp.}
	\label{fig:ae2}
\end{figure}

Figure \ref{fig:ae1} shows the variation of speech-based emotion over the video length where ``Happy'', ``Surprised'', ``Neutral'', ``Fear'' are dominant. To generate this figure, we extracted 10 secs of speech with a moving window of 10 secs and a hop of 10 secs as well. Subsequently, the model prediction probability for every emotion was used for plotting. Similarly, Figure \ref{fig:ae2} shows the variation of speech-based emotion over the other video where ``Anger'', ``Sad'', ``Disgust'', ``Neutral'' are dominant.

\section{Discussions \& Conclusion}
We have developed a new framework CLUE for contextualised adaptive engagement. Some previous works, e.g., \cite{bulathwela2020predicting} have focused on the textual information only for establishing contextualised engagement which is insufficient to provide feedback. As a result, we have extracted additional discriminative features, which are, textual emotion, speech emotion variability with time, animation and object detection from the video lectures, and unified them to create a prediction variable and update the vector based on user feedback. Our novel model unifies the individual pre-trained models and learns their weight parameters in a completely unsupervised way. Our results show that our model can reliably provide engagement scores followed by explainability which existing models cannot do.

The individual models in our CLUE framework were first trained on publicly available datasets, and the training performances were reported on those datasets as baseline results. Subsequently, we unified these models to develop our novel CLUE framework and compared the performance of each model against the CLUE model. As evidenced in the results in Figure~\ref{fig:score} the major impact on the prediction is based on $X_1$ which was trained on the VLN dataset with subset features, used in \cite{bulathwela2020predicting}. Removing $X_2$, from the overall model, did not impact the prediction significantly which models the emotions based on the textual content. Likely, textual emotions are not very crucial in domain-specific videos. As a result, the emotion of text can have the least impact on domain-specific predictions. Removing $X_3$, which is based on emotion decoding over speech reduced the predicted engagement score significantly. It is also observed that variation of positive emotion increases engagement compared to negative emotion. Figure \ref{fig:ae1} shows the variation in emotion of speech over time where ``Happy'', ``Surprised'', ``Neutral'', ``Fear'' are dominant and Figure \ref{fig:ae2} shows the variation in emotion of speech over time where ``Anger'', ``Sad'', ``Disgust'', ``Neutral'' are dominant. Engagement score of video, Figure \ref{fig:ae1}, were significantly better than the video in Figure \ref{fig:ae2}. Variation in the emotion of speech over time helps to increase the engagement score than having a single emotion tone for a longer period. Removing $X_4$ which accounts for object count over the video also had a significant drop in the engagement score as the animation plays a key role in engagement. However, the impact of this model would be greater if we could account for more than 80 objects as fixed in Yolo. Based on textual data the length of the video, tobe\_verb\_rate,  Auxilary\_rate, speaker speed, and the easiness level of the text is important for engagement prediction. For video to be engaging the length of the video should be short, the use of tobe\_verb\_rate should be $<0.03\%$, use of Auxilary\_rate should be $<0.025\%$, speaker speed should be in the range of 115-120 wpm, easiness level should be more than 83.

In the future, we will design an estimator and policy based on the observed truth $y$ and ground truth $\hat{y}$ to make the prediction stronger and user-centric. To this end, we will develop a new reinforcement learning framework where user feedback is incorporated into the model.
\bibliographystyle{ACM-Reference-Format}
\bibliography{WSDM}


%\newpage
\appendix

\section{Real-world study of our contextualised engagement model after deployment in a production environment by Learning Management System (LMS) company}
Our model was deployed online in a production environment and was provided for public use. LMS organisation helped in assessing online lectures and evaluating them in real-time with their users. The deployed model evaluated the two sample lectures from NPTEL, one a lecture on Friction and the other on eVehicles, based on the design and delivery of the content. The design of the lecture was evaluated on Impact, Complexity, Content Richness and Segmentation whereas the delivery was evaluated based on Vividness, Facial Expression, Speech Expression, and Speech Speed Performance. The goal of this user study is to measure whether our model can learn reliably from the data. The data used in this study can be obtained by contacting the authors which not only includes the videos but also the associated user study data.

Our model was deployed on the server and the LMS company asked tutors to upload their online content on the server, where the model evaluated every video and provided feedback to the creators. Post analysis LMS company asked students to mark the video on a scale of 1 to 10 in terms of engagement. Post lecture, content creators asked questions related to the topic in form of a checkbox to gauge the understanding of content.

\begin{enumerate}
    \item Impact -- Content impact measures the proportion of meaningful words to that of total words used in the content. The briefer the content is the higher is the learner engagement. Impact measure helps maintain brevity by creating maximum impact through the lowest possible content.
    \item Complexity -- Content complexity measures the depth and diversity of the words used in design and delivery. Content complexity feedback helps optimise the usage of rare and unique words to keep the content simple and thus maximise learner engagement. 
    \item Content Richness -- Content richness provides feedback on optimal usage of educational media in the video. A video lecture gets the highest possible visual engagement when it has more multimedia in it. However, creating such a video is complex and highly time-consuming. Content richness helps instructors identify the optimal amount of educational media to be used without worrying about the engagement level of the video.
    \item Segmentation -- Segmentation provides feedback on the optimal duration of the video for better learner attention. 
    \item Vividness -- Content vividness measures the contextual word usage in speaker communication. The usage of striking words is proven to better engage listeners over bland word choices. Appropriate word choices eliminate the overuse of words and reduce the monotonicity in the narration.
    \item Facial Expression -- Speaker facial decoding reports the emotions expressed in the speaker's face during content delivery. Major emotions that are reported and proven to impact the learner mood are happy, sad, surprise, angry, disgust and fear. 
    \item Speech Expression -- Delivery tone and rhythm feedback report the emotions conveyed in the speaker's audio pitch. Speech tone and rhythm are found to influence the listener's mental state. The feedback helps the speaker achieve a tone that expresses joy and surprise that is proven to have better engagement among the listeners.
    \item Speech Speed Performance -- Speaker speed measures the words-per-minute in speech. Speaker speed influences the processing of the learner's acquired information. While a lower speed is proven to be perceived as less challenging among the listeners, speed in the upper range is proven to be equally dangerous in losing learner attention. Speaker speed feedback helps optimize the speed based on the context of the content.
\end{enumerate}

\begin{figure}
	\centering
	\includegraphics[width=0.43\textwidth,keepaspectratio]{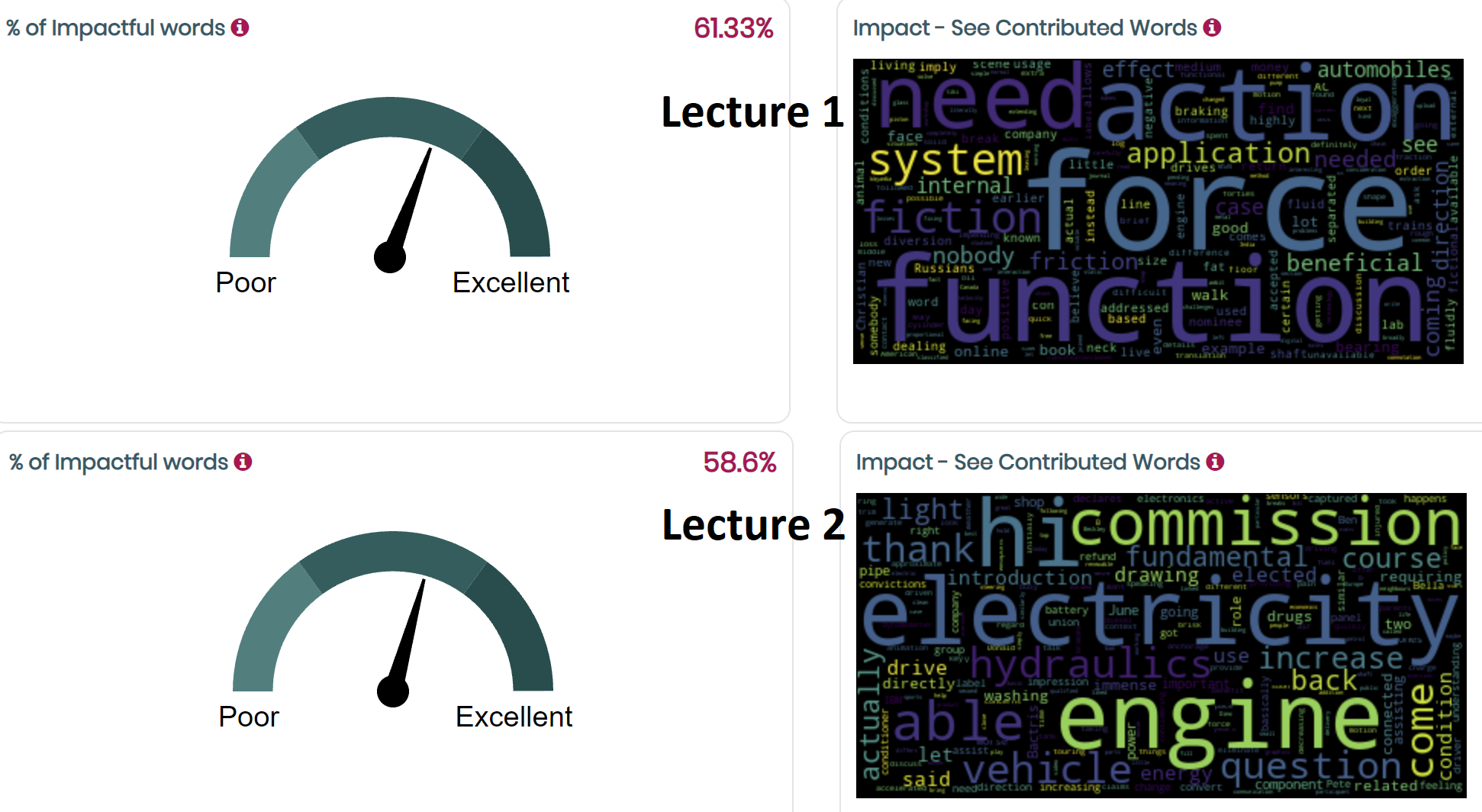}
	\caption{The two lectures which were measured in their of the impact that they can create.}
	\label{fig:imp}
\end{figure}

\begin{figure}
	\centering
	\includegraphics[width=0.43\textwidth,keepaspectratio]{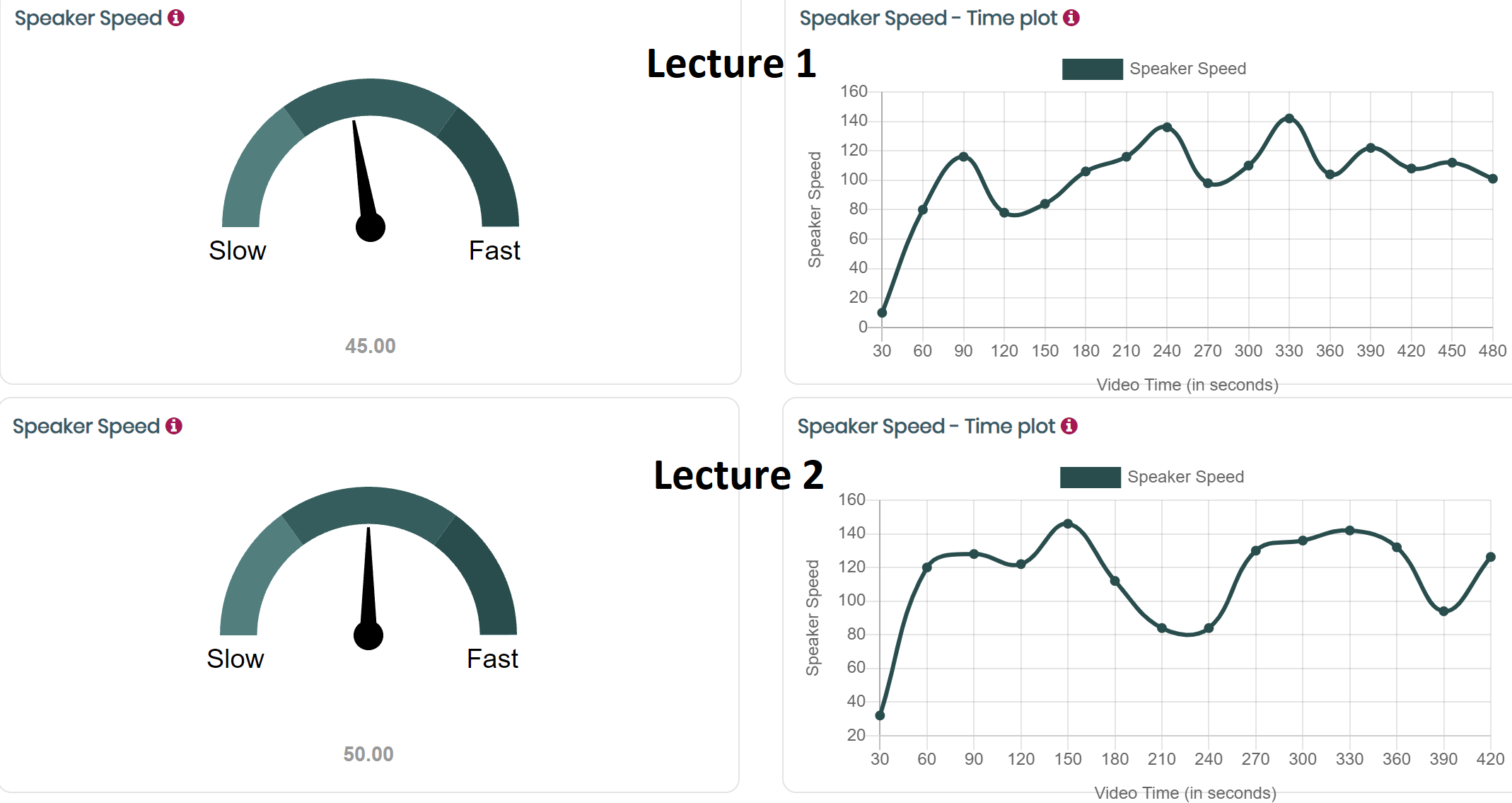}
	\caption{Comparison between two videos on the basis of speaker speed.}
	\label{fig:speed}
\end{figure}

\begin{figure}
	\centering
	\includegraphics[width=0.43\textwidth,keepaspectratio]{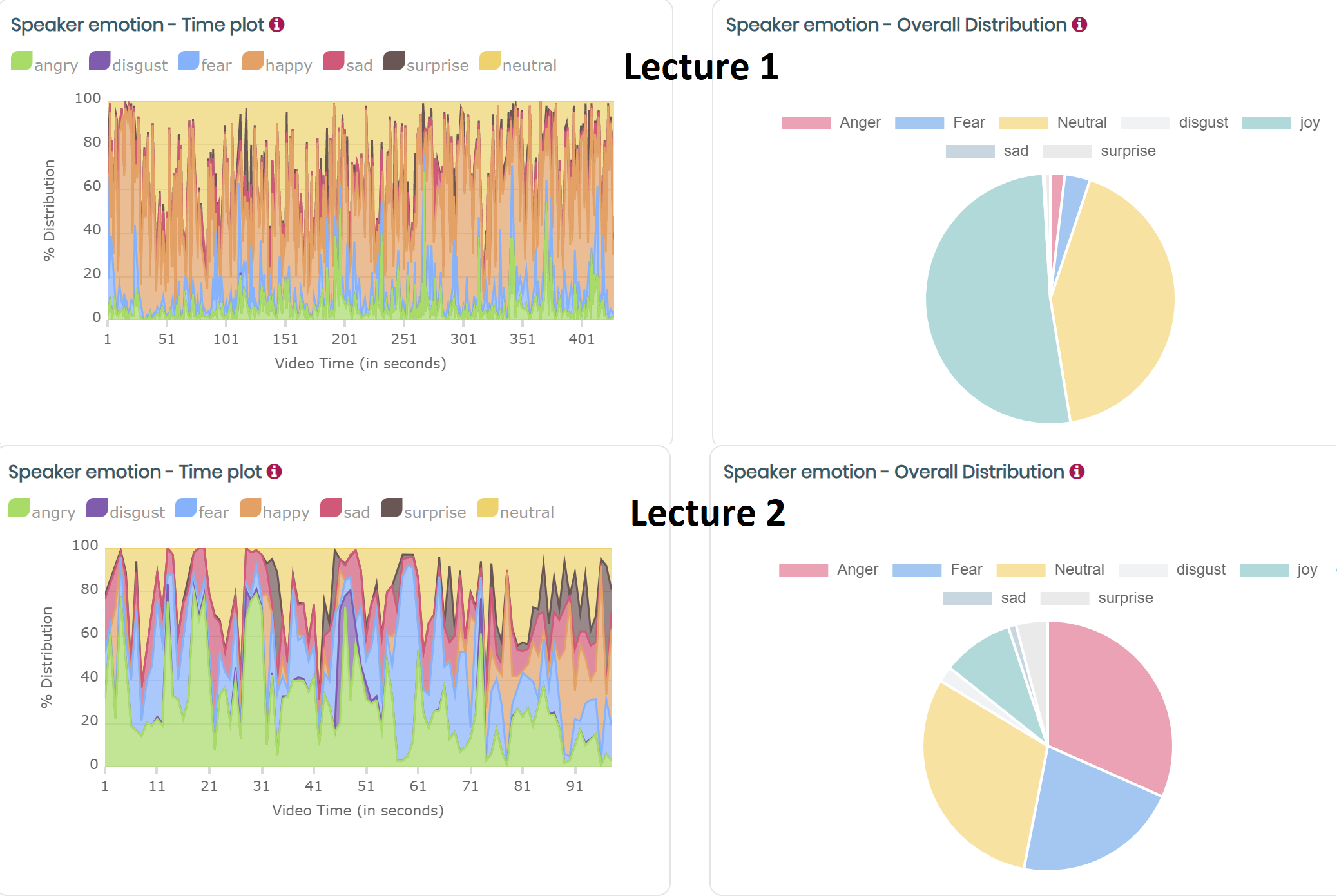}
	\caption{Comparison between two videos on the basis of speaker emotion.}
	\label{fig:emotion}
\end{figure}

\begin{figure}
	\centering
	\includegraphics[width=0.43\textwidth,keepaspectratio]{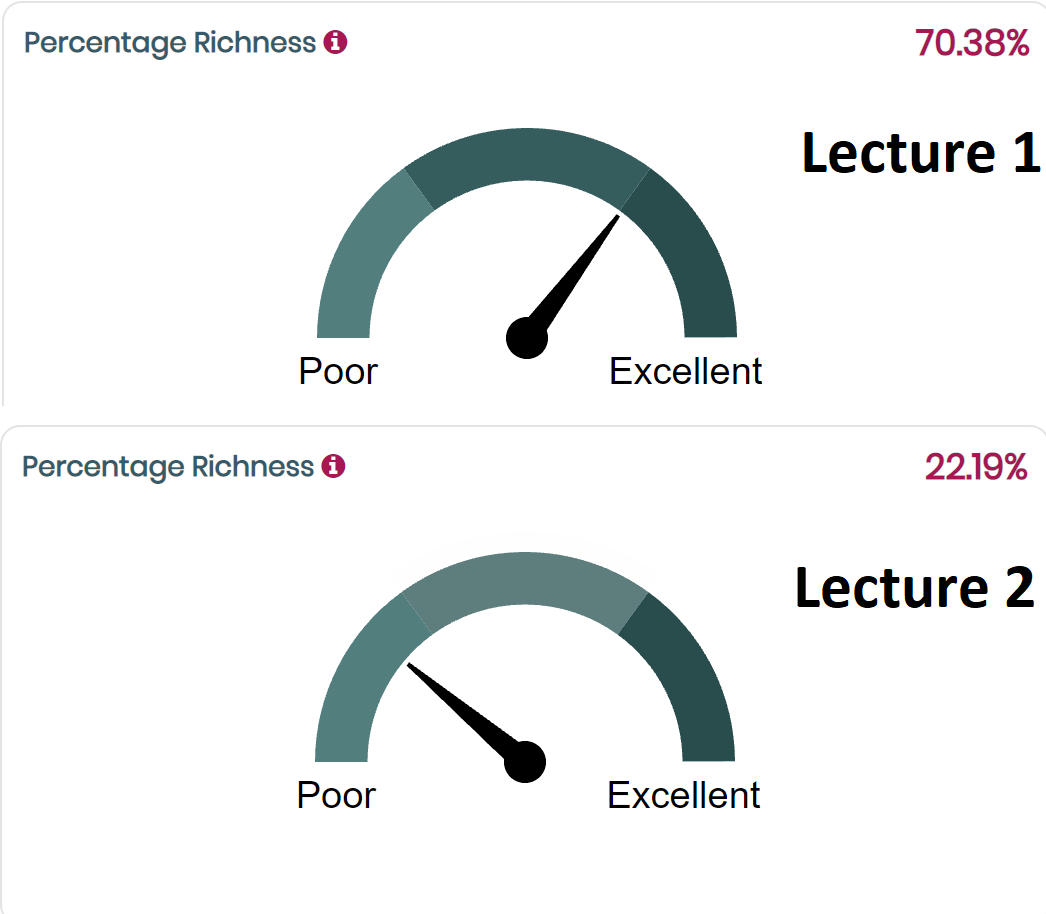}
	\caption{Comparison between two videos on the basis of richness of content.}
	\label{fig:richness}
\end{figure}

\begin{figure*}
	\centering
	\includegraphics[width=0.8\textwidth,keepaspectratio]{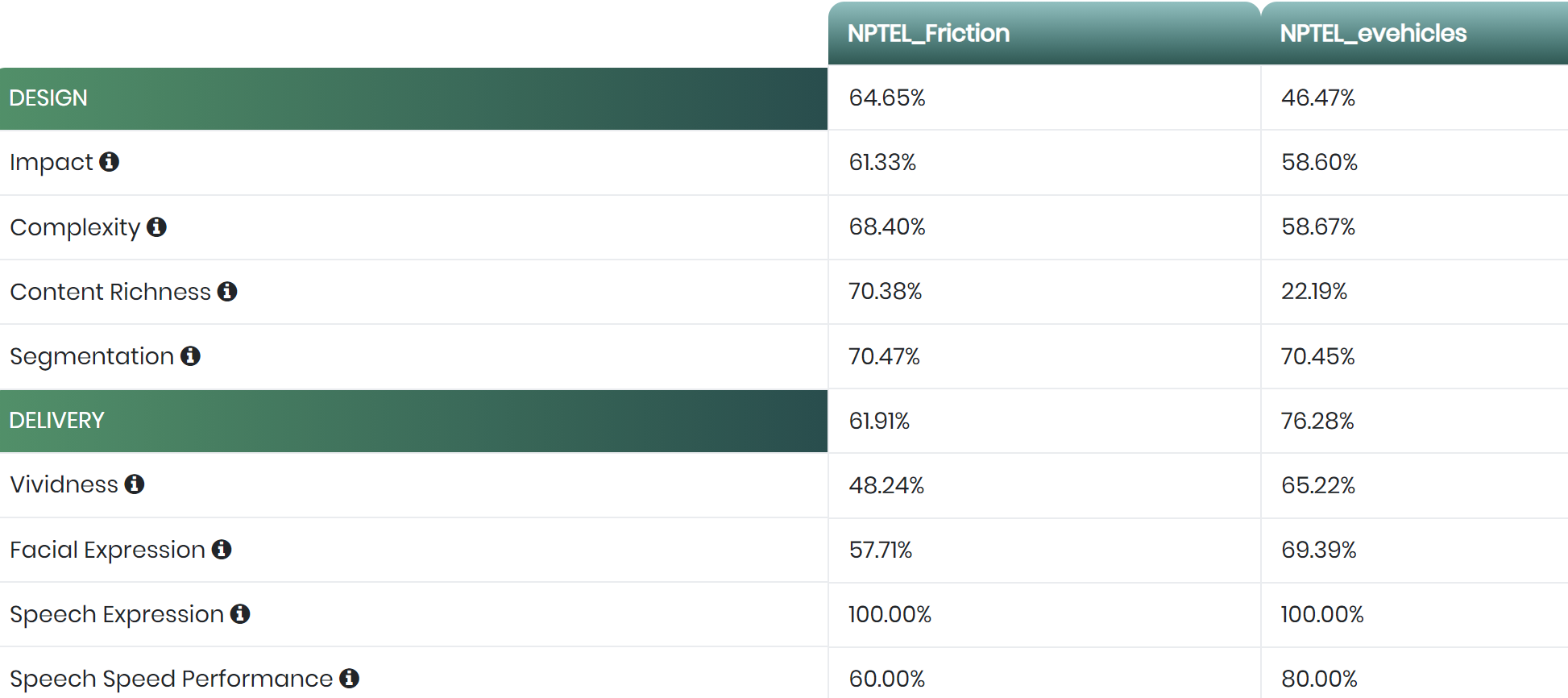}
	\caption{Comparison between two videos content on the basis of design and delivery.}
	\label{fig:comp}
\end{figure*}
\end{document}